\documentclass[10pt,twocolumn,letterpaper]{article}

\usepackage{cvpr}              \usepackage{bbding}

\usepackage[dvipsnames]{xcolor}

\definecolor{cvprblue}{rgb}{0.21,0.49,0.74}
\usepackage[pagebackref,breaklinks,colorlinks,citecolor=cvprblue]
{hyperref}

\title{Can LLM-Generated Text Empower Surgical Vision-Language Pre-training?}

\author{
Chengan Che\thanks{Co-first authors, equal contribution.}\quad
Chao Wang\footnotemark[1]\quad
Jiayuan Huang\quad
Xinyue Chen\quad
Luis C. Garcia-Peraza-Herrera \\
\vspace{0.5pt} \\
Visual Understanding Research Group, Department of Informatics, King’s College London, UK \\
}

\usepackage{booktabs}
\usepackage{graphicx}
\usepackage{amsmath}
\usepackage{multirow} % For multi-row cells
\usepackage{tabularx}  % Automatically adjust table width
\usepackage{listings}
\usepackage{hyperref}
\usepackage{subcaption}
\usepackage{pifont} 

\usepackage[table]{xcolor}

\usepackage{xcolor}
\usepackage{algorithm}
\usepackage{algpseudocode}

\definecolor{MyGreen}{HTML}{E2EFDA}      % A specific "Sea Green"
\definecolor{MyBlue}{HTML}{DDEBF7} 
\definecolor{videoblue}{HTML}{4E85D9} % A dark teal
\definecolor{imageorange}{HTML}{E97133} % A dark teal
\usepackage{placeins}

\usepackage{caption}

\makeatletter
\let\@oldmaketitle\@maketitle
\renewcommand{\@maketitle}{\@oldmaketitle
  \begin{center}
    \vspace{-1.5em} % Pulls the image closer to the author block if needed
    \includegraphics[width=\linewidth]{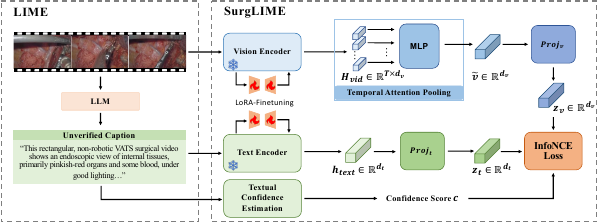}
    \captionof{figure}{\textbf{Overview of our proposed LIME dataset and SurgLIME framework.} We first employ a Large Language Model (Gemini) to generate narratives for surgical video clips from LEMON~\cite{Che2026LEMON}, establishing the LIME dataset(Sec.~\ref{sec:LIME}). Within the SurgLIME architecture (Sec.~\ref{sec:method}), a frozen vision encoder (PL-Stitch~\cite{che2026stitch}) equipped with LoRA~\cite{Hu2021LoRA:Models} extracts frame-level embeddings $H_{\text{vid}}$. These are dynamically aggregated by a Temporal Attention Pooling module into a unified video representation $\tilde{v}$. In parallel, the text encoder (PubMedBERT~\cite{Gu2021Domain-SpecificProcessing}) extracts the textual embedding $h_{text}$, while an automated scoring mechanism computes a textual confidence score $c_i$. Modality-specific projection heads then map these representations ($\tilde{v}$ and $h_{text}$) into a shared metric space, yielding $z_v$ and $z_t$. Finally, $c_i$ acts as a soft weight to dynamically modulate the bidirectional InfoNCE contrastive loss~\cite{DBLP:journals/corr/abs-1807-03748}, explicitly down-weighting the influence of hallucinated pseudo-labels when computing the final objective $\mathcal{L}_{total}$.}
    \label{fig:overview}
  \end{center}
  \vspace{1em} % Adds space between the figure and the abstract
}
\makeatother

\begin{document}
\maketitle
\begin{abstract}

Recent advancements in self-supervised learning have led to powerful surgical vision encoders capable of spatiotemporal understanding. 
However, extending these visual foundations to multi-modal reasoning tasks is severely bottlenecked by the prohibitive cost of expert textual annotations. 
To overcome this scalability limitation, we introduce \textbf{LIME}, a large-scale multi-modal dataset derived from open-access surgical videos using human-free, Large Language Model (LLM)-generated narratives. 
While LIME offers immense scalability, unverified generated texts may contain errors, including hallucinations, that could potentially lead to catastrophically degraded pre-trained medical priors in standard contrastive pipelines. 
To mitigate this, we propose \textbf{SurgLIME}, a parameter-efficient Vision-Language Pre-training (VLP) framework designed to learn reliable cross-modal alignments using noisy narratives.
SurgLIME preserves foundational medical priors using a LoRA-adapted dual-encoder architecture and introduces an automated confidence estimation mechanism that dynamically down-weights uncertain text during contrastive alignment. 
Evaluations on the AutoLaparo and Cholec80 benchmarks show that SurgLIME achieves competitive zero-shot cross-modal alignment while preserving the robust linear probing performance of the visual foundation model.
Dataset, code, and models are publicly available at \href{https://github.com/visurg-ai/SurgLIME}{https://github.com/visurg-ai/SurgLIME}.

\end{abstract}

\section{Introduction}
\label{sec:intro}
Recent advancements in self-supervised learning have yielded remarkably powerful surgical vision encoders~\cite{che2026stitch, Wang2023FoundationPre-train, Che2026LEMON, Batic2024EndoViT:Images, Jaspers2026ScalingModels}. 
These foundational models encode profound medical priors and extract robust visual representations of operative scenes.
However, they are limited to the visual modality. 
To unlock advanced, open-vocabulary reasoning tasks, such as surgical question answering or zero-shot phase recognition, a semantic ``bridge'' is required to connect these rich visual embeddings with textual descriptions. 
However, acquiring high-quality surgical text requires extensive curation and verification by medical experts~\cite{Meireles2021SAGESVideo, NyangohTimoh2023AVideo, Ward2021ChallengesAnnotation}. 
This severe scalability bottleneck significantly hinders the development of surgical vision-language models.

This problem motivates a highly practical question: \textit{Can we reduce reliance on human experts by utilizing Large Language Model (LLM)-generated narratives to establish this cross-modal bridge?}
While LLM-generated medical texts are scalable, they introduce a severe secondary challenge: they are inherently noisy and prone to critical hallucinations~\cite{zhu-etal-2025-trust, 10.1145/3571730}.
Standard Vision-Language Pretraining (VLP) architectures typically learn joint visual and textual representations through contrastive objectives (e.g., InfoNCE~\cite{DBLP:journals/corr/abs-1807-03748}), which implicitly assume reliable video–text correspondences. 
When trained with noisy LLM-generated narratives, incorrect or hallucinated descriptions can corrupt the contrastive signal, leading to misaligned representations and unstable optimization.

To mitigate this issue, we hypothesize that the robust medical priors encoded in a pre-trained surgical vision encoder can serve as a stabilizing anchor. 
Rather than fully finetuning the visual and textual encoders under noisy supervision, we preserve the pre-trained representations to leverage the strong visual manifold established by PL-Stitch~\cite{che2026stitch}. 
We investigate whether this approach can guide the alignment process, aiming to learn robust cross-modal representations despite imperfect textual supervision.
%
% As illustrated in Fig.~\ref{fig:overview}, we explore this hypothesis by first introducing \textbf{LIME}, an \textbf{L}LM-\textbf{I}nferred \textbf{M}ultimodal \textbf{E}ndoscopy dataset derived from the open-access LEMON archive~\cite{Che2026LEMON}. To align visual and textual modalities while preserving the integrity of the pre-trained surgical vision encoder, we propose the \textbf{SurgLIME} framework.

As illustrated in Fig.~\ref{fig:overview}, we explore this hypothesis by first introducing \textbf{LIME}, an \textbf{L}LM-\textbf{I}nferred \textbf{M}ultimodal \textbf{E}ndoscopy dataset (Sec.~\ref{sec:LIME}) derived from the open-access LEMON dataset~\cite{Che2026LEMON}. 
To align visual and textual modalities, we design an exploratory framework, \textbf{SurgLIME} (Sec.~\ref{sec:method}).
Unlike standard VLP pipelines~\cite{Li2022BLIP:Generation, Radford2021LearningSupervision} that treat text supervision as reliable, SurgLIME operates under the assumption that the text is inherently flawed. 
As a first step towards solving this, we adopt a dual parameter-efficient fine-tuning strategy. 
We freeze both the self-supervised surgical vision foundation (PL-Stitch \cite{che2026stitch}) and the pre-trained text encoder (PubMedBERT~\cite{Gu2021Domain-SpecificProcessing}), injecting Low-Rank Adaptation (LoRA)~\cite{Hu2021LoRA:Models} modules into both streams to align the modalities without disrupting their foundational priors.
Furthermore, we introduce a PubMedBERT-driven~\cite{Gu2021Domain-SpecificProcessing} confidence weighting scheme to dynamically down-weight hallucinated text during the contrastive alignment process.

Evaluations on standard benchmarks indicate that this approach achieves viable cross-modal alignment while maintaining the integrity of the visual foundation. 
We provide open access to our dataset, models, and code.

Our contributions are summarized as follows:
\begin{itemize}
    \item We introduce the \textbf{LIME} dataset, exploring the viability of using unverified, human-free generated text to bridge the modality gap in surgical vision-language learning.
    
    \item We propose \textbf{SurgLIME}, a parameter-efficient VLP framework that integrates LoRA-adapted foundational encoders and a dynamic textual confidence weighting mechanism to learn cross-modal representations from noisy narratives.
    
    \item Evaluations on Cholec80~\cite{Twinanda2017EndoNet:Videos} and AutoLaparo~\cite{Wang2022AutoLaparo:Hysterectomy} indicate that SurgLIME yields viable zero-shot cross-modal alignment and preserves the semantic richness of the pre-trained visual manifold, as reflected by its robust linear probing performance.
\end{itemize}

% Evaluations on standard benchmarks, including Cholec80~\cite{Twinanda2017EndoNet:Videos} and AutoLaparo~\cite{Wang2022AutoLaparo:Hysterectomy}, suggest that SurgLIME can achieve viable cross-modal alignment without degrading the semantic richness of the pre-trained visual manifold.
% Our contributions are summarized as follows:
% \begin{itemize}
%     \item We introduce the \textbf{LIME} dataset, exploring the viability of using unverified, human-free generated text to bridge the modality gap in surgical vision-language learning.
    
%     \item We propose \textbf{SurgLIME}, a parameter-efficient VLP framework that integrates LoRA-adapted foundational encoders and a dynamic textual confidence weighting mechanism to learn robust cross-modal representations from noisy narratives.
    
%     \item Experiments indicate that SurgLIME yields viable zero-shot cross-modal alignment and preserves linear probing performance, supporting the feasibility of scaling surgical VLP via LLM-generated noisy narratives.
% \end{itemize}

\begin{figure*}[!t]
  \centering
  \includegraphics[width=.97\linewidth]{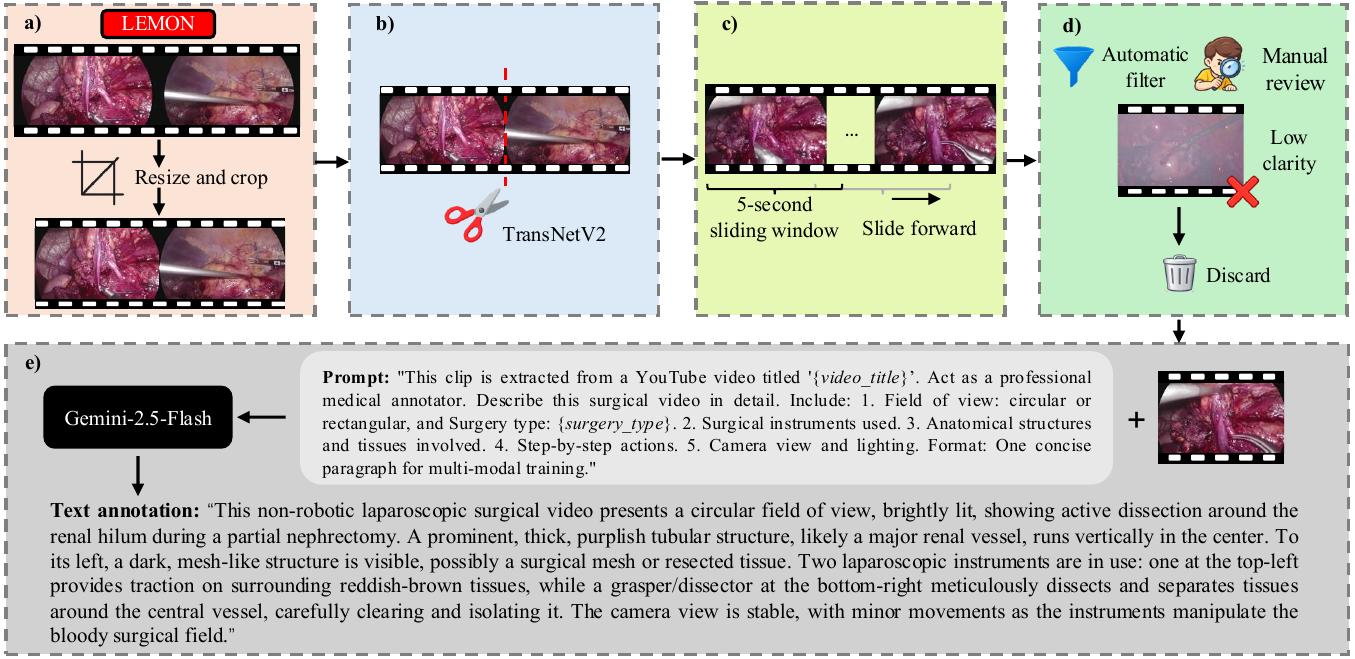}
  \caption{\textbf{Overview of the LIME dataset construction pipeline.} \textbf{a)} The process begins with  standardizing raw videos from the LEMON dataset through resizing and center-cropping. \textbf{b)} Long-form videos are then partitioned using TransNetV2 for shot boundary detection, followed by \textbf{c)} a 5-second sliding window approach to generate temporal segments. \textbf{d)} High-clarity clips are selected via an automatic Laplacian-based filter, followed by a manual review to remove residual blurred clips. \textbf{e)} Finally, remaining clips are paired with detailed textual annotations generated by Gemini-2.5-Flash using a structured prompt, resulting in a multi-modal dataset for downstream training.}
  \label{fig:scheme}
\end{figure*}

\section{Related Work}
\label{sec:related_work}

\noindent
\textbf{Surgical datasets.}
The generalization capabilities of general-domain vision-language (VL) foundation models are fundamentally driven by massive, web-sourced image-text datasets \cite{Sharma2018ConceptualCaptioning, Schuhmann2022LAION-5B:Models}. 
This immense scale of data has empowered architectures such as CLIP~\cite{Radford2021LearningSupervision} and BLIP \cite{Li2022BLIP:Generation} to excel across a wide spectrum of tasks, ranging from zero-shot retrieval to complex spatial and logical reasoning \cite{Xie2026SpatiaLQA:Models, Gong2025Med-CMR:Reasoning}, adaptive self-correction \cite{Zhang2026NotReasoning}, and multi-modal anchoring \cite{Zhang2026Chain-of-ThoughtAnchoring}. 
Adapting this open-vocabulary success to the surgical domain, however, is severely bottlenecked by privacy regulations and the prohibitive cost of expert clinical annotation~\cite{Meireles2021SAGESVideo, NyangohTimoh2023AVideo, Ward2021ChallengesAnnotation}. 
Conventional surgical datasets \cite{Twinanda2017EndoNet:Videos, Wang2022AutoLaparo:Hysterectomy, Che2026LEMON, Jaspers2026ScalingModels} are confined to the visual modality, providing either unannotated video or closed-set labels. 
This absence of textual descriptions restricts models to narrow, predefined tasks.
To enable flexible multi-modal reasoning, recent efforts have introduced surgical video-text datasets, such as SurgLaVi \cite{Perez2026SurgLaVi:Learning}, which pairs surgical clips with clinical descriptions. 
Despite this progress, scaling medically accurate annotations remains resource-intensive. 
To address this, we augment LEMON dataset \cite{Che2026LEMON} with LLM-generated captions. 
This serves to empirically investigate whether unverified, noisy supervision can provide meaningful utility for surgical vision-language pre-training.

% \noindent
% \textbf{Self-supervised pre-training.}
% %
% Recent self-supervised learning (SSL) \cite{Caron2021EmergingTransformers, Caron2020UnsupervisedAssignments, Chen2020ImprovedLearning, Su2025StreamlineDistillation, Chen2020ARepresentations, Du2026UnsupervisedDecoupling, Wang2023VideoMAEMasking, Assran2025V-JEPAPlanning, che2026stitch} has established robust feature foundations by circumventing manual annotations. Extending beyond purely visual representations, Vision-Language Pre-training (VLP)~\cite{Radford2021LearningSupervision, Li2022BLIP:Generation} learns a shared metric space that aligns visual semantics with rich textual contexts. By mapping these modalities together through large-scale contrastive learning, VLP enables powerful cross-modal capabilities~\cite{Xie2026SpatiaLQA:Models, Li2021AlignDistillation, Jia2021ScalingSupervision}. However, applying general VLP architectures directly to surgery yields sub-optimal results due to specialized domain vocabulary, visually homogeneous anatomies, and complex procedural workflows. To address this, models like SurgVLP \cite{Yuan2025LearningLectures} adapt contrastive objectives to align surgical frames with expert-annotated clinical text. 
% %
% In this work, we inject low-rank adapters \cite{Hu2021LoRA:Models} and temporal pooling into a robust visual foundation model (PL-Stitch~\cite{che2026stitch}) to test whether noisy, LLM-generated text can yield meaningful cross-modal alignments without degrading pre-trained representations.

\noindent
\textbf{Self-supervised pre-training.}
Recent self-supervised learning (SSL) \cite{Caron2021EmergingTransformers, Caron2020UnsupervisedAssignments, Chen2020ImprovedLearning, Su2025StreamlineDistillation, Chen2020ARepresentations, Du2026UnsupervisedDecoupling, Wang2023VideoMAEMasking, Assran2025V-JEPAPlanning, che2026stitch} has established robust feature foundations by circumventing manual annotations.
Extending beyond purely visual representations, Vision-Language Pre-training (VLP)~\cite{Radford2021LearningSupervision, Li2022BLIP:Generation} learns a shared metric space that aligns visual semantics with rich textual contexts. 
By mapping these modalities together through large-scale contrastive learning, VLP enables powerful cross-modal capabilities~\cite{Xie2026SpatiaLQA:Models, Li2021AlignDistillation, Jia2021ScalingSupervision, Su_2026_WACV, su2025medgrpo}. 
However, applying general VLP models directly to surgery yields sub-optimal results due to specialized domain vocabulary, visually homogeneous anatomies, and complex procedural workflows. 
To address this, models like SurgVLP \cite{Yuan2025LearningLectures} adapt contrastive objectives to align surgical frames with transcribed audio narrations from clinical lectures.
However, this approach severely limits scalability due to its inherent reliance on scarce, expert-narrated lecture videos.

In this work, we seek a scalable alternative to expert annotations and constrained ASR transcripts of surgical lectures. 
Inspired by the noise-robustness of general-domain VLP \cite{Jia2021ScalingSupervision}, we investigate whether contrastive learning can handle the algorithmic noise inherent to LLM-generated captions, which are structurally coherent yet potentially hallucinated. 
Specifically, we inject low-rank adapters \cite{Hu2021LoRA:Models} and temporal pooling into a robust visual foundation model (PL-Stitch~\cite{che2026stitch}) to test whether such unverified supervision can still yield meaningful cross-modal alignments without degrading pre-trained representations.
%
% However, applying general VLP architectures directly to surgery yields sub-optimal results due to specialized domain vocabulary, visually homogeneous anatomies, and complex procedural workflows. To address this, models like SurgVLP \cite{Yuan2025LearningLectures} adapt contrastive objectives to align surgical frames with transcribed audio narrations from clinical lectures.

% While both unscalable expert annotations and lecture-constrained ASR transcripts offer domain-specific supervision, we explore a more universally scalable alternative: LLM-generated pseudo-captions. Inspired by the noise-robustness of contrastive objectives demonstrated in general-domain VLP \cite{Jia2021ScalingSupervision}, we investigate the utility of this approach. These captions present a unique "high-grammar, low-fact" algorithmic noise---structurally coherent but potentially hallucinated. In this work, we inject low-rank adapters \cite{Hu2021LoRA:Models} and temporal pooling into a robust visual foundation model (PL-Stitch~\cite{che2026stitch}) to test whether such unverified supervision can still yield meaningful cross-modal alignments without degrading pre-trained representations.

\section{Proposed Dataset: LIME}
\label{sec:LIME}

To facilitate the training of multi-modal models specialized for the surgical domain, we curated a high-quality video-text dataset LIME derived from the LEMON~\cite{Che2026LEMON} dataset, the largest open-source repository of surgical videos to date. 
The original LEMON collection comprises 4194 surgical videos sourced from YouTube, with durations ranging from several minutes to nearly an hour. 
As shown in Fig.~\ref{fig:scheme}, we developed a multi-stage automated pipeline to transform these raw videos into a collection of 54k surgical clips with dense semantic annotations.

\noindent
\textbf{Resolution standardization.} 
We first standardized the raw videos to ensure computational consistency, we performed a shortest-side resize followed by a center crop to achieve a fixed resolution of $832 \times 480$ pixels.

\noindent
\textbf{Shot segmentation. }
To ensure semantic coherence within each clip, we employed TransNetV2~\cite{soucek2024transnet} for automated shot boundary detection. 
Long-form videos were partitioned into discrete shots based on detected transitions. 
We discarded any shots shorter than 5 seconds, as such brief intervals typically lack sufficient temporal context.

\noindent
\textbf{Temporal standardization. }
To further unify the input format for model training, we applied a sliding window approach. 
Each segment was decomposed into clips of approximately 5 seconds. 
This duration is chosen to be sufficient to capture a meaningful interaction between instruments and tissues, while remaining within the optimal comprehension range of current multi-modal LLMs.
We utilized a window size of 5 seconds with a stride of 2 seconds, which improves sample diversity via various temporal offsets while simultaneously mitigating excessive information redundancy between overlapping clips.

\noindent
\textbf{Data pruning. }
To prevent inaccurate multi-modal LLM descriptions caused by degraded inputs, we eliminated blurred clips by first applying a Laplacian sharpness filter, followed by a manual review.

\noindent
\textbf{Automated captioning with multi-modal LLM.}
Finally, we leveraged Gemini-2.5-Flash~\cite{team2023gemini} to generate detailed, domain-specific linguistic descriptions for the remaining clips. 
To maximize the comprehensiveness of the annotations, we utilized a structured prompt incorporating original video metadata (e.g., video title and surgery type). 
The prompt instructed the model to act as a professional medical annotator, focusing on: (1) Field of view (circular vs. rectangular) and surgery type (robotic vs. non-robotic)~\cite{Che2026LEMON}; (2) Surgical instruments utilized; (3) Anatomical structures and involved tissues; (4) Step-by-step actions and procedural maneuvers; (5) Camera perspective and lighting conditions. The final output was formatted into a single, concise paragraph, resulting in a densely captioned surgical dataset prepared for training multi-modal models.

\section{Proposed Framework: SurgLIME}
\label{sec:method}

In this section, we detail the proposed SurgLIME framework, as illustrated in Fig.~\ref{fig:overview}. 
We introduce a parameter efficient dual encoder architecture and a confidence weighted contrastive objective designed to learn robust representations from noisy LLM generated text.
% To learn robust cross-modal representations from noisy LLM-generated pseudo-supervision, we propose a parameter-efficient Vision-Language Pre-training (VLP) framework. 
% %
% As illustrated in Fig.~\ref{fig:overview}, our approach utilizes a dual-encoder equipped with temporal attention pooling to extract unified multi-modal embeddings. 
% %
% To align these visual and textual representations, we optimize a contrastive objective (i.e., InfoNCE) that is weighted by an automated textual confidence score, effectively filtering generated hallucinations during training.

\subsection{Problem Formulation}
\label{sec:problem_formulation}

Our primary objective is to learn a robust surgical vision-language representation from a dataset of LLM-generated video-text pairs. Formally, we define the training dataset as 
$\mathcal{D} = \{(V_i, S_i, c_i)\}_{i=1}^N$, 
where each sample consists of a surgical video clip $V_i$, a corresponding generated textual sentence $S_i$, and an explicitly derived confidence score $c_i \in (0, 1]$ indicating the estimated reliability of the textual descriptions.
Unlike static medical imaging, surgical phases are continuous, context-dependent macroscopic events. Therefore, we define the input visual modality as a temporal sequence of $T$ frames, $V_i = \{v_1, v_2, \dots, v_T\}$, where each frame $v_t \in \mathbb{R}^{H \times W \times C}$. 

Our goal is to optimize two modality-specific mappings: a vision branch $f_v$ and a text branch $f_t$. 
The vision branch processes the temporal window to extract a unified, video-level visual embedding $z_v \in \mathbb{R}^D$. Concurrently, the text branch maps the surgical description into a textual embedding $z_t \in \mathbb{R}^D$ within the same shared metric space. 
By aligning $z_v$ and $z_t$ through a noise-aware contrastive objective modulated by $c_i$, we aim to obtain a highly generalized visual foundation model capable of zero-shot transfer and robust linear probing, despite the inherent hallucination risks in the generated text.

\subsection{Parameter-Efficient Dual Encoders}
\label{sec:encoders}

Given the noisy nature of $\mathcal{D}$, we avoid fully fine-tuning the encoders to mitigate the risk of the model overfitting to textual hallucinations and compromising the pre-trained visual representations.
To bridge the modality gap, we freeze the pre-trained weights of both encoders and inject Low-Rank Adaptation (LoRA) modules~\cite{Hu2021LoRA:Models} into their attention layers.

\noindent
\textbf{Vision encoder.} We utilize \textbf{PL-Stitch}~\cite{che2026stitch}, a surgical vision foundation model (ViT-Base) robustly pre-trained on the large-scale surgical dataset LEMON~\cite{Che2026LEMON}, as our visual foundation. 
The backbone weights are frozen to preserve its generalized dense prediction capabilities.
Following standard PEFT practices~\cite{Hu2021LoRA:Models}, low-rank trainable matrices are injected into the query, key, and value (QKV) projection matrices of all self-attention blocks. 
For an input video clip $V_i$, the spatial encoder processes the $T$ frames independently to extract frame-level global embeddings, yielding a sequence of visual features $H_{vid} = \{h_1, h_2, \dots, h_T\}$, where $h_t \in \mathbb{R}^{d_v}$.

\noindent
\textbf{Text encoder.} For the textual domain, we employ \textbf{PubMedBERT}~\cite{Gu2021Domain-SpecificProcessing}, a domain-specific model pre-trained on biomedical corpora, to accurately extract surgical semantics from the narratives.
Similarly, its base parameters are frozen, and LoRA modules are injected into the query and value matrices. 
Given a tokenized surgical description $S_i$, the text encoder outputs a global sentence representation $h_{text} \in \mathbb{R}^{d_t}$ via its \texttt{[CLS]} token.

\subsection{Temporal Attention Pooling}
\label{sec:temporal_pooling}

Surgical phase recognition requires extended temporal context. 
To generate a visual clip representation, a naive approach would be to apply static mean pooling on the individual frame embeddings, treating all frames equally.
However, this could potentially render the representation vulnerable to sudden occlusions (e.g., smoke, blood) and abrupt camera motions. 
To dynamically aggregate the frame-level representations $H_{vid}$ into a unified semantic embedding, we introduce a learnable Temporal Attention Pooling module.
%
% Surgical phases require macroscopic temporal context.
% %
% To generate a visual clip representation, a naive approach would be to apply static mean pooling to treat all frames equally. 
% %
% However, this renders the representation vulnerable to sudden occlusions (e.g., smoke, blood) and abrupt camera motions. 
% %
% To dynamically aggregate the frame-level representations $H_{vid}$ into a unified semantic embedding, we introduce a learnable Temporal Attention Pooling module.
% %
% Static mean pooling treats all frames equally, rendering the representation vulnerable to sudden occlusions (e.g., smoke, blood) or abrupt camera motions.

The module computes a scalar attention weight for each frame $h_t$ using a two-layer Multi-Layer Perceptron (MLP) with a $\tanh$ bottleneck, which is then normalized across the temporal dimension $T$ via a softmax function to produce the final video-level representation $\tilde{v}$:
\begin{align}
    s_t &= W_2 \tanh(W_1 h_t), \nonumber \\
    a_t &= \frac{\exp(s_t)}{\sum_{j=1}^T \exp(s_j)}, \quad \tilde{v} = \sum_{t=1}^T a_t h_t,
    \label{eq:temporal_pooling}
\end{align}
where $W_1 \in \mathbb{R}^{\frac{d_v}{2} \times d_v}$ and $W_2 \in \mathbb{R}^{1 \times \frac{d_v}{2}}$ are trainable weights, and $\tilde{v} \in \mathbb{R}^{d_v}$ encapsulates the temporally smoothed, macro-level visual state of the surgical clip.

\subsection{Textual Confidence Estimation}
\label{sec:confidence}

To explicitly mitigate the hallucination risks inherent in LIME, we introduce a confidence scoring mechanism to quantify the reliability of each LLM-generated narrative. 
We leverage PubMedBERT~\cite{Gu2021Domain-SpecificProcessing} to perform a token prediction evaluation. 
Formally, given a generated narrative $S_i$ consisting of $L_i$ tokens, $S_i = \{w_1, w_2, \dots, w_{L_i}\}$, we iteratively replace each token $w_k$ with a \texttt{[MASK]} token to construct a masked context $S_{i \backslash k}$. 
The final confidence score $c_i \in (0, 1]$ is defined as the average probability of recovering the original tokens using PubMedBERT:
\begin{equation}
    c_i = \frac{1}{L_i} \sum_{k=1}^{L_i} P_{\text{MB}}(w_k \mid S_{i \backslash k}).
    \label{eq:confidence}
\end{equation}
$P_{\text{MB}}$ denotes the softmax-normalized predicted probability from the masked language modeling head based on the surrounding context. Consequently, sentences with high linguistic and medical plausibility yield higher average recovery probabilities, while highly uncertain or hallucinated descriptions are automatically assigned lower scores.

\subsection{Confidence-Weighted Cross-Modal Alignment}
\label{sec:objective}

To align the temporally aggregated visual feature $\tilde{v}$ and the textual feature $h_{text}$, we map them into a shared $D$-dimensional metric space using modality-specific projection heads. Each projector ($\text{Proj}_v$ and $\text{Proj}_t$) consists of a two-layer MLP combined with Layer Normalization and a GELU activation. 
%
%To satisfy the constraints of the contrastive objective, the projected features are strictly L2-normalized:
Following standard practice~\cite{Chen2020ARepresentations, Radford2021LearningSupervision} in contrastive learning, the projected features are strictly L2-normalized to map them onto a unit hypersphere:
\begin{equation}
    z_v = \frac{\text{Proj}_v(\tilde{v})}{\|\text{Proj}_v(\tilde{v})\|_2}, 
    \quad 
    z_t = \frac{\text{Proj}_t(h_{text})}{\|\text{Proj}_t(h_{text})\|_2}.
    \label{eq:normalization}
\end{equation}

We optimize the network using a bidirectional InfoNCE contrastive loss~\cite{DBLP:journals/corr/abs-1807-03748} dynamically modulated by our derived confidence scores. 
Given a batch of $B$ video-text pairs, the confidence score $c_i$ acts as a weight to explicitly penalize the loss contribution of uncertain LLM-generated narratives.
The visual-to-text loss $\mathcal{L}_{v \to t}^{(i)}$ for the $i$-th sample, alongside the final averaged bidirectional objective $\mathcal{L}_{total}$, are formulated as:
\begin{align}
    \mathcal{L}_{v \to t}^{(i)} &= - c_i \log \frac{\exp(z_{v}^{(i)} \cdot z_{t}^{(i)} / \tau)}{\sum_{j=1}^B \exp(z_{v}^{(i)} \cdot z_{t}^{(j)} / \tau)}, \\
    \mathcal{L}_{total} &= \frac{1}{2B} \sum_{i=1}^B \left( \mathcal{L}_{v \to t}^{(i)} + \mathcal{L}_{t \to v}^{(i)} \right),
    \label{eq:loss}
\end{align}

% \begin{align}
%     \mathcal{L}_{v \to t}^{(i)} &= - c_i \log \frac{\exp\left( \frac{z_{v}^{(i)} \cdot z_{t}^{(i)}}{\tau} \right)}{\sum_{j=1}^B \exp\left( \frac{z_{v}^{(i)} \cdot z_{t}^{(j)}}{\tau} \right)}, \\
%     \mathcal{L}_{total} &= \frac{1}{2B} \sum_{i=1}^B \left( \mathcal{L}_{v \to t}^{(i)} + \mathcal{L}_{t \to v}^{(i)} \right),
%     \label{eq:loss}
% \end{align}
where $\tau$ is a learnable temperature parameter. The symmetric text-to-visual loss $\mathcal{L}_{t \to v}^{(i)}$ is computed identically over the transposed similarity matrix. 

By decoupling the learning rates, specifically applying a higher multiplier to the randomly initialized projectors and pooling layer while maintaining a low base rate for the LoRA weights, we ensure stable convergence without disrupting the pre-trained manifolds.

\section{Experiments}
\label{sec:experiments}

In this section, we first detail the experimental setup in Sec.~\ref{sec:setup}. Next, we present quantitative comparisons for surgical phase recognition via zero-shot evaluation in Sec.~\ref{sec:exp_zeroshot} and linear probing in Sec.~\ref{sec:exp_linear}. Finally, we analyze the impact of textual confidence estimation in Sec.~\ref{sec:exp_ablation}.

\begin{table}[!t]
    \centering
    \small
    \setlength{\tabcolsep}{1.9pt}
    \caption{
        \textbf{Zero-shot surgical phase recognition results.} 
        We report video-level accuracy and F1-score for zero-shot evaluations on the AutoLaparo and Cholec80 datasets.
        The experiments are conducted by computing the cosine similarity between the visual embeddings and the textual embeddings of prompt-augmented phase descriptions, without any supervised fine-tuning on the target datasets.
        Best in \textbf{bold}.
    }
    \label{tab:zeroshot}
    \begin{tabularx}{\columnwidth}{@{}lccccccc@{}}
    \hline
        \multirow{2}{*}{Method} &\multirow{2}{*}{Backbone} &~ & \multicolumn{2}{c}{AutoLaparo} & ~ & \multicolumn{2}{c}{Cholec80} \\
        \cline{4-5} \cline{7-8} 
        & & & Acc & F1-score & & Acc & F1-score \\
        \hline

        CLIP~\cite{Radford2021LearningSupervision} & ViT-B/16 & ~
        & $8.0$ & $4.8$ & ~ 
        & $27.8$ & $8.4$ 
        \\

        SurgVLP~\cite{Yuan2025LearningLectures} & ResNet50 & ~
        & $10.0$ & $7.2$ & ~ 
        & $\mathbf{34.7}$ & $\mathbf{24.4}$ 
        \\

        \textbf{SurgLIME (ours)} & ViT-B/16 & ~
        & $\mathbf{18.1}$ & $\mathbf{11.2}$ & ~ 
        & 33.0 & 8.7 
        \\
        \hline
    \end{tabularx}
\end{table}

\begin{table*}[t!]
    \centering
    \small
    \setlength{\tabcolsep}{7.4pt}
    \caption{
        \textbf{Linear probing results.}
        We report top-1 accuracy and F1-score on the AutoLaparo and Cholec80 datasets. The experiments are conducted with a frozen visual backbone to verify that our cross-modal alignment preserves the integrity of the pre-trained visual representations. All predictions are computed on a frame-by-frame basis. Type `S' denotes a surgical-specific foundation model, `G' denotes a generalist visual self-supervised model, and `VL' denotes a vision-language pre-trained model. Best in \textbf{bold}.
        }
    \label{tab:linear_probing_classification}
    \begin{tabularx}{.72\linewidth}{@{}lcccccccc@{}}
    \hline
        \multirow{2}{*}{Method}  &\multirow{2}{*}{Type} &\multirow{2}{*}{Backbone} &~ & \multicolumn{2}{c}{AutoLaparo} & ~ & \multicolumn{2}{c}{Cholec80} \\
        \cline{5-6} \cline{8-9} 
        & & & & Acc & F1-score & & Acc & F1-score \\
        \hline

        % Endo-FM~\cite{Wang2023FoundationPre-train} &  & ViT-B/16 & ~
        % & $51.5$ & $43.1$ & ~ 
        % & ~ 
        % & $62.7$  & $53.9$ 

        LemonFM~\cite{Che2026LEMON}  &\multirow{1}{*}{S} & ConvNeXt-B & ~
        & $74.7$ & $64.5$ & ~ 
        & $73.9$ & $65.8$ 
        \\
        \hline

        MAE~\cite{He2022MaskedLearners} &\multirow{5}{*}{G} & ViT-B/16 &  ~
        & $35.5$ & $32.0$ & ~ 
        & $54.9$ & $43.4$ 
         \\  

        VideoMAEv2~\cite{Wang2023VideoMAEMasking} & ~ & ViT-B/16 & ~
        & $49.8$ & $42.4$ & ~ 
        & $55.8$ & $48.5$ 
         \\  

        DINO~\cite{Caron2021EmergingTransformers} & ~ & ViT-B/16 & ~
        & $74.9$ & $65.0$ & ~ 
        & $72.2$ & $67.1$ 
        \\

        iBOT~\cite{Zhou2022IBOT:Tokenizer}  & ~ & ViT-B/16 & ~
        & $76.3$ & $65.1$ & ~ 
        & $74.6$ & $67.6$ 
        \\

        PL-Stitch~\cite{che2026stitch} & ~ & ViT-B/16 & ~
        & 79.9 & $\mathbf{69.0}$ & ~ 
        & 80.4 & 73.0 
        \\
        \hline

        CLIP~\cite{Radford2021LearningSupervision} & \multirow{3}{*}{VL} & ViT-B/16 & ~
        & $53.1$ & $42.1$ & ~ 
        & $64.8$ & $50.7$ 
        \\

        SurgVLP~\cite{Yuan2025LearningLectures} & ~ & ResNet50 & ~
        & $54.3$ & $41.8$ & ~ 
        & $63.5$ & $50.3$ 
        \\

        \textbf{SurgLIME (ours)} & ~ & ViT-B/16 & ~
        & $\mathbf{80.7}$ & 68.8 & ~ 
        & $\mathbf{80.6}$ & $\mathbf{73.2}$ 
        \\
        \hline
    \end{tabularx}
\end{table*}

\subsection{Experimental Setup}
\label{sec:setup}

\textbf{Datasets and evaluation protocols.} 
We evaluate our model on surgical phase recognition task using two widely recognized surgical benchmarks: 
(1) \textbf{Cholec80} \cite{Twinanda2017EndoNet:Videos}, which contains 80 cholecystectomy videos categorized into seven surgical phases; 
and (2) \textbf{AutoLaparo} \cite{Wang2022AutoLaparo:Hysterectomy}, which consists of 21 laparoscopic hysterectomy videos with seven defined phases, providing a challenging domain for temporal reasoning.
To rigorously assess cross-modal capabilities and feature quality, we utilize two protocols:
(1) \textbf{Zero-shot Evaluation}: This task directly evaluates the model's cross-modal semantic alignment. 
Specifically, we utilize the frozen text encoder to generate embeddings by passing detailed descriptions of each surgical phase, using the prompting templates defined in~\cite{Perez2026SurgLaVi:Learning}.
For a given test video clip, the video-level representation is extracted via the vision encoder and temporal attention pooler. The model predicts the phase by identifying the textual embedding with the highest cosine similarity to the video embedding. 
Crucially, no task-specific training or fine-tuning is performed on the target datasets.
Consistent with prior works~\cite{Perez2026SurgLaVi:Learning, Yuan2025LearningLectures}, we report video-level Accuracy and F1-score as the primary metrics.
(2) \textbf{Linear Probing Evaluation}: To verify if the noisy VLP supervision has compromised the visual foundation, we freeze the vision encoder and train a linear classifier using the official train/test splits.
This evaluates the discriminative quality and utility of the visual features after they have been aligned with noisy narratives. We report frame-wise Accuracy and F1-score following~\cite{Che2026LEMON, che2026stitch}.

\noindent
\textbf{Implementation details.} 
SurgLIME is implemented using the PyTorch framework. The vision encoder is a ViT-Base initialized with PL-Stitch weights~\cite{che2026stitch}, while the text encoder is based on PubMedBERT~\cite{Gu2021Domain-SpecificProcessing}. We inject LoRA modules~\cite{Hu2021LoRA:Models} into both encoders with a rank $r=16$ and a scaling factor $\alpha=32$. The model is pre-trained on the proposed \textbf{LIME} dataset for 10 epochs using the AdamW optimizer with a base learning rate of $2\times 10^{-4}$ and a cosine decay schedule. To capture macroscopic surgical events, a temporal window of $T=8$ frames is processed via the learnable attention-based pooling layer.

\subsection{Zero-shot Evaluation}
\label{sec:exp_zeroshot}
Operating under the zero-shot protocol, SurgLIME relies entirely on the semantic alignment synthesized from the noisy, LLM-generated Lemon dataset. As summarized in Table~\ref{tab:zeroshot}, SurgLIME consistently outperforms the standard CLIP \cite{Radford2021LearningSupervision} baseline across both benchmarks. 
Notably, on AutoLaparo, SurgLIME outperforms the recent state-of-the-art SurgVLP~\cite{Yuan2025LearningLectures} by an absolute margin of \textbf{8.1pp} in accuracy.
On the Cholec80 benchmark, although SurgVLP achieves the highest zero-shot accuracy by leveraging transcribed expert lectures that explicitly detail standardized workflows, SurgLIME remains competitive and continues to surpass the CLIP model.
These results underscore the viability of our framework, demonstrating that LLM-generated narratives can serve as a cross-modal bridge to align surgical visual features with textual semantics.

\subsection{Linear Probing Evaluation}
\label{sec:exp_linear}

To conduct the linear probing evaluation, we first merge the learned LoRA weights back into the frozen ViT backbone prior to training the linear classifier. 

The results, detailed in Table \ref{tab:linear_probing_classification}, indicate that SurgLIME successfully preserves the strong discriminative power of the PL-Stitch \cite{che2026stitch} foundation. Notably, on the AutoLaparo benchmark, the proposed framework yields a slight accuracy increase of $\approx$1 percentage point over the PL-Stitch baseline. 
This demonstrates that our noise-aware cross-modal pre-training not only avoids catastrophic forgetting but also induces a marginal shift in the visual manifold. Ultimately, these findings confirm that parameter-efficient fine-tuning allows the model to acquire new modality-alignment capabilities without degrading its pre-existing surgical visual foundation.

\subsection{Ablation on Textual Confidence Estimation}
\label{sec:exp_ablation}

To isolate the impact of textual confidence weighting ($c_i$), we evaluate a standard unweighted InfoNCE baseline ($c_i=1$). On the AutoLaparo zero-shot task, this baseline achieves 13.4\% accuracy and 10.1\% F1. By dynamically penalizing uncertain generated narratives, SurgLIME improves this to \textbf{18.1\%} and \textbf{11.2\%}, respectively. This confirms that explicitly down-weighting unverified LLM hallucinations is critical to prevent cross-modal feature degradation.

\section{Conclusion}
\label{sec:conclusion}
In this paper, we explore the viability of surgical vision-language pre-training using unverified, human-free generated text.
We introduce LIME, a scalable multi-modal dataset, and propose SurgLIME, a parameter-efficient framework that employs a confidence-weighted contrastive objective to learn cross-modal representations using LLM-generated narratives.
Experimental results confirm that this approach establishes competitive zero-shot phase recognition and maintains the discriminative integrity of the visual foundation. 
Ultimately, this work provides a baseline for developing multi-modal surgical models capable of learning from noisy generated text without relying on prohibitive expert annotations or constrained ASR transcripts of surgical lectures.
Future work will focus on methodological enhancements, such as developing finer-grained, token-level confidence weighting mechanisms and iterative self-correction protocols for LLM-generated supervision, alongside extending the framework to complex downstream tasks like surgical captioning and robotic action generation.

\FloatBarrier

{
    \small
    \bibliographystyle{ieeenat_fullname}
    \bibliography{main}
}

\end{document}